\documentclass[letterpaper, 10 pt, conference]{ieeeconf}  

\IEEEoverridecommandlockouts                              

\usepackage{graphics} 
\usepackage{epsfig} 
\usepackage{mathptmx} 
\usepackage{times} 
\usepackage{amsmath} 
\usepackage{amssymb}  

\usepackage{graphicx}
\usepackage{booktabs}
\usepackage{dsfont}

\usepackage{bbding}
\usepackage{pifont}
\usepackage{wasysym}
\usepackage{color, colortbl}

\usepackage{algorithm}
\usepackage{algpseudocode}
\usepackage{stackengine}
\usepackage{multirow}
\usepackage{tabularx}
\usepackage{subcaption}
\usepackage{esvect}
\usepackage{makecell}
\usepackage{overpic}
\usepackage{wrapfig}
\usepackage{tikz, xcolor}
\usetikzlibrary{calc,spy}

\usepackage[pagebackref,breaklinks,colorlinks]{hyperref}
\usepackage[capitalize]{cleveref}

\title{\LARGE \bf
BroadBEV: Collaborative LiDAR-camera Fusion \\
for Broad-sighted Bird's Eye View Map Construction
}

\author{Minsu Kim$^{1\dagger}$, Giseop Kim$^1$, Kyong Hwan Jin$^{2}$, Sunwook Choi$^{1*}$%
\thanks{$^*$Corresponding author.}%
\thanks{$\dagger$ Work done during an internship at NAVER LABS.}%
\thanks{$^{1}$Minsu Kim, Giseop Kim, and Sunwook Choi are with Autonomous Driving Group, NAVER LABS {\tt\small \{minshu.kim, giseop.kim, sunwook.choi\}@naverlabs.com}
}%
\thanks{$^{2}$Kyong Hwan Jin is with School of Electrical Engineering, Korea University, Seoul, Republic of Korea {\tt\small kyong\_jin@korea.ac.kr}}
}
\begin{document}

\maketitle
\thispagestyle{empty}
\pagestyle{empty}

\begin{abstract}
A recent sensor fusion in a Bird's Eye View (BEV) space has shown its utility in various tasks such as 3D detection, map segmentation, etc. However, the approach struggles with inaccurate camera BEV estimation, and a perception of distant areas due to the sparsity of LiDAR points. In this paper, we propose a broad BEV fusion (\textit{BroadBEV}) that addresses the problems with a spatial synchronization approach of cross-modality. Our strategy aims to enhance camera BEV estimation for a broad-sighted perception while simultaneously improving the completion of LiDAR's sparsity in the entire BEV space. Toward that end, we devise \textit{Point-scattering} that scatters LiDAR BEV distribution to camera depth distribution. The method boosts the learning of depth estimation of the camera branch and induces accurate location of dense camera features in BEV space. For an effective BEV fusion between the spatially synchronized features, we suggest \textit{ColFusion} that applies self-attention weights of LiDAR and camera BEV features to each other. Our extensive experiments demonstrate that BroadBEV provides a broad-sighted BEV perception with remarkable performance gains.
\end{abstract}


\section{Introduction}
Visual perception and understanding of the surrounding environment are crucial to implementing reliable robotic systems such as Simultaneous Localization and Mapping (SLAM), and Advanced Driver Assistance Systems (ADAS). Because the perception provides an ego-frame agent with detailed local features and structural information, various approaches to the perceptions have been actively studied including 3D detection and semantic segmentation. As a representation of latent variables for the tasks, Bird's Eye View (BEV) space has been frequently employed. BEV space is free from distortions of homogeneous coordinate systems and categorizes object shapes into a few classes. Thus, it provides a robust representation of elements in 3D space including cars, buildings, pedestrians, large-scale scenes, etc.

Recently, approaches using the fusion of multiple sensors' features in a shared BEV space \cite{liang2022bevfusion, liu2023bevfusion, borse2023x, li2022hdmapnet} have demonstrated effective representations even for images and points with noise and motion blur. Because each sensor has clearly distinct strong and weak points, complementary usages between sensors lead to the achievement of consistent and robust model performances. For example, in LiDAR-camera fusion, LiDAR detects accurate depth values of surrounding environments but has sparse scene contexts in far regions. In contrast, the camera obtains dense 2D signals projected on a camera plane but loses corresponding depth information. Thus, an optimal fusion can be summarized as a method that improves a camera branch's depth learning and LiDAR branch's sparsity completion.

\begin{figure}[t]
    \begin{subfigure}[b]{\linewidth}
        \centering
        \includegraphics[width=0.90\linewidth]{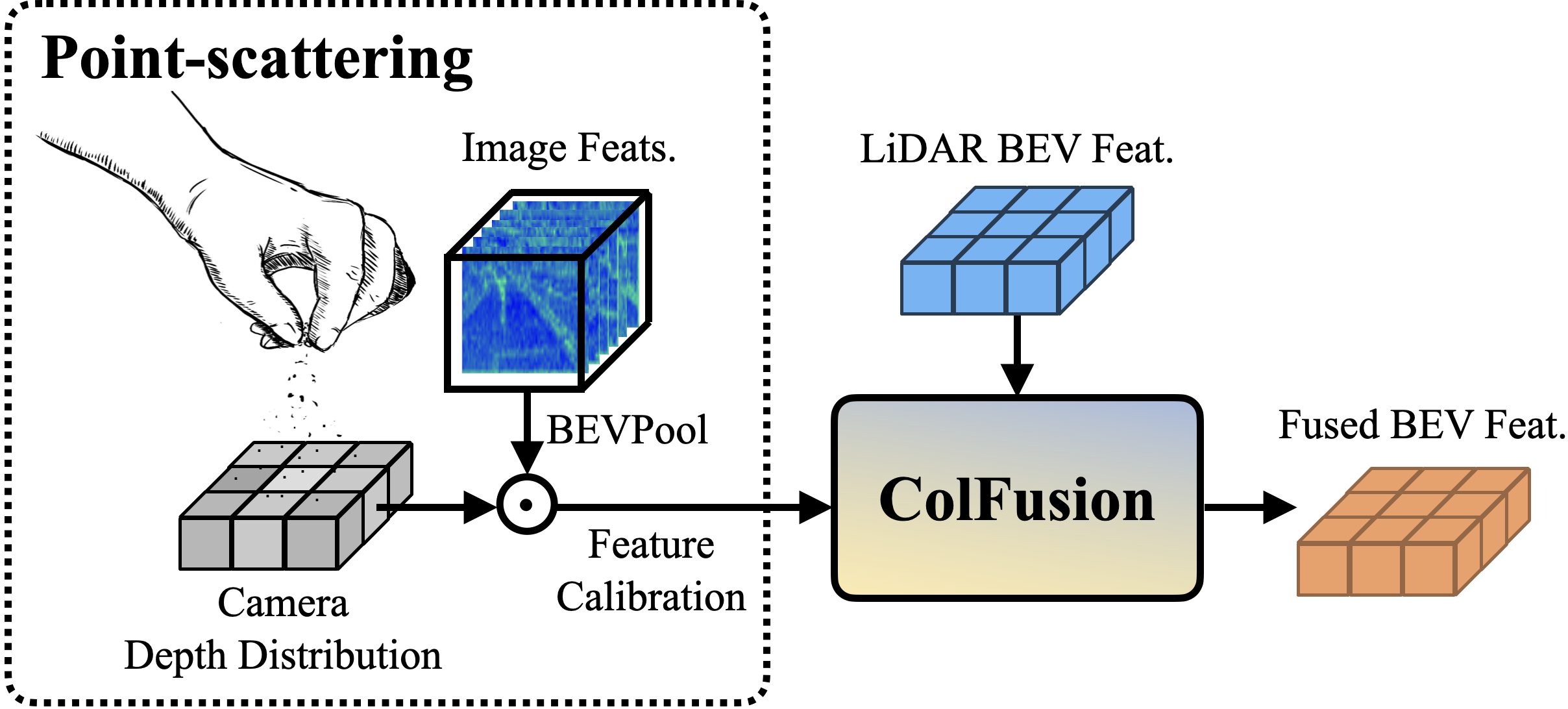}
        \subcaption{Overview of BroadBEV.}
        \vspace{4pt}
    \end{subfigure}
    \begin{subfigure}[b]{\linewidth}
        \centering
        \includegraphics[width=0.90\linewidth]{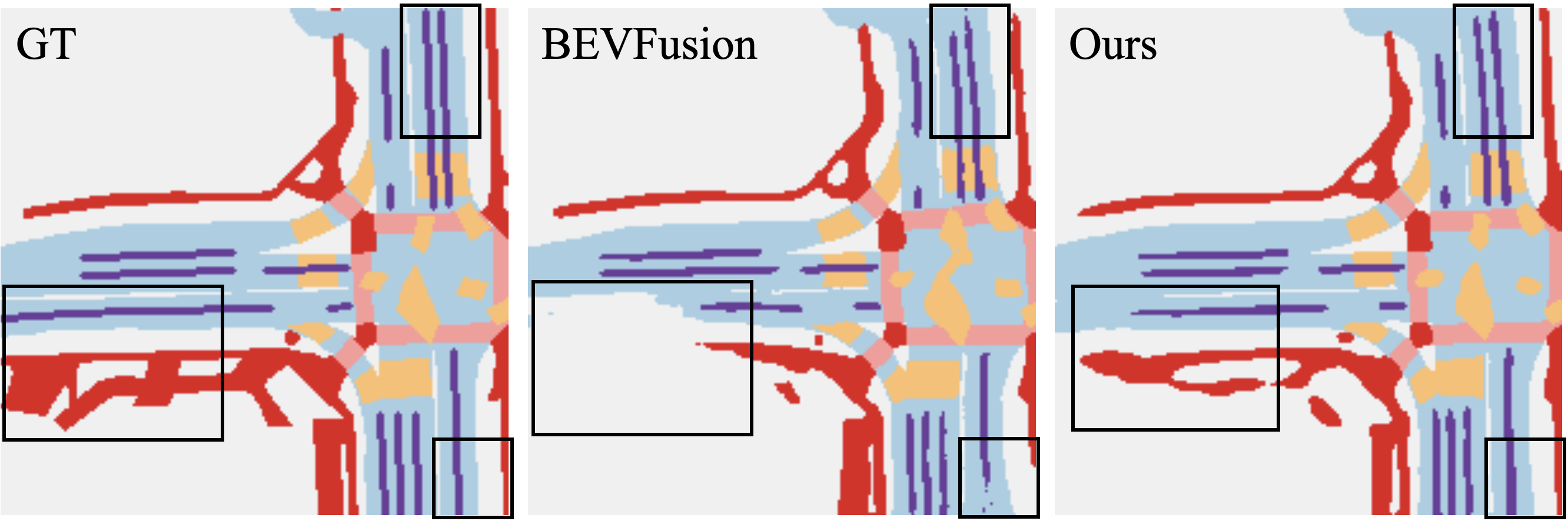}
        \subcaption{A Comparison of Broadness on Map Segmentation Results.}
    \end{subfigure}
    \vspace{-14pt}
    \caption{BroadBEV scatters LiDAR points to camera depth distribution to guarantee geometric synchronization of cross-modality. After it evaluates a camera BEV feature, Our collaborative fusion (or ColFusion) ensembles the BEV features of LiDAR and cameras. As shown in the bottom comparison, our method provides a broad-sighted perception.}
    \label{fig:teaser}
    \vspace{-18pt}
\end{figure}

The existing implementations of BEV fusion usually lift and splat \cite{philion2020lift} images to transform 2D features into 3D space. The view transform is practical as it prevents BEV projection from long-tail artifacts caused by Inverse Perspective Mapping (IPM) \cite{hartley2003multiple}. However, because the lifting needs image depth distribution, it is inevitable to use monocular depth estimation. As a result, the requirement sometimes imposes inaccurate depths on a camera branch and negatively affects the performance. The limitation becomes a bottleneck for a model's perception of distant regions because it will fail to reasonably interpolate the large sparsity of LiDAR due to incorrect camera depths.

\begin{figure*}[t]
    \centering
    \includegraphics[width=0.85\linewidth]{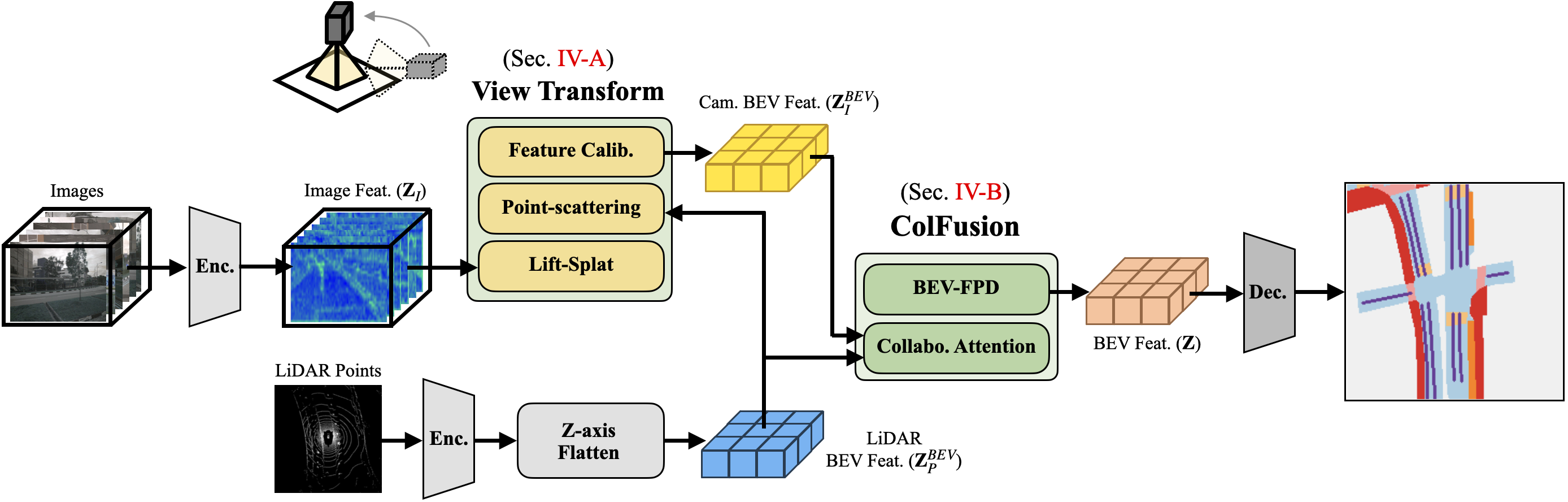}
    \caption{\textbf{Overall Scheme of BroadBEV.} Our view transform takes in LiDAR BEV feature ($\mathbf{Z}^{BEV}_P$) to lift and splat \cite{philion2020lift} image features ($\mathbf{Z_I}$) with our proposed Point-scattering to camera depth distribution. Then it calibrates the camera BEV feature. After preparing camera and LiDAR BEV features, ColFusion which consists of collaborative attention and BEV-FPD \cite{wang2023lidar2map} fuses them into a unified BEV feature ($\mathbf{Z}$). BroadBEV decodes the feature into a constructed map.}
    \label{fig:flowchart}
    \vspace{-16pt}
\end{figure*}

To address the problem, we suggest a broad BEV fusion (BroadBEV). Specifically, it synchronizes the 3D geometry of sensors in BEV space to share LiDAR's consistent sensing to a camera branch. To implement the idea, we devise a \textit{Point-scattering} that scatters LiDAR BEV distribution to the estimated camera depth distribution. Based on the enhanced camera BEV features, our collaborative fusion (\textit{ColFusion} in short) applies attention weights of LiDAR and camera BEV features to each other for boosting model robustness. 

Our BroadBEV provides a broad-sighted BEV perception because our enhancement of the camera branch induces the improved completion of LiDAR's large sparsity in distant areas. We evaluate our approach to semantic map construction as the task requires the accurate positioning of context features with precise depths in a BEV space. The extensive experiments in \cref{sec:exp} show that our method achieves the best performance. As far as we know, our BroadBEV is the first approach to explore 
a broad-sighted BEV fusion. Our contributions are summarized as:
\begin{itemize}
\setlength{\leftmargin}{-.35in}
    \item Our Point-scattering guarantees a synchronized geometry to the camera and LiDAR BEV branches. Our experiments in \cref{sec:exp} validate that it contributes to the extraction of an enhanced camera BEV feature.
    \item For an effective fusion of BEV features with shared 3D geometry, we propose a novel collaborative fusion (or ColFusion). In our ablation in \cref{tab:ablation}, this method boosts model performance.
    \item BroadBEV provides a broad-sighted perception. With the strong point, our approach shows the best performances in extensive experiments.
\end{itemize}

\section{Related Works}
\noindent\textbf{Camera BEV Representation.} To encode perspective images to BEV features, various view transformation methods have been employed including Inverse Perspective Mapping (IPM) \cite{reiher2020sim2real, kim2019deep, zhu2021monocular, qiao2023end}, Lift-Splat \cite{philion2020lift} based explicit transformation \cite{huang2021bevdet, huang2022bevdet4d, roddick2018orthographic, hong2022cross, li2023bevdepth}, and query-based implicit transformation \cite{palazzi2017learning, carion2020end, 
xie2022m, li2022bevformer, yang2023bevformer, pan2020cross, shin2023instagram}. Because the methods have to estimate depth or learn an implicit BEV space with monocular camera features, they show weak generalization in challenging environments such as rainy days, and low-light conditions. To be free from the problem, fusion with LiDAR is able to be adopted. Because the sensor is relatively independent of environments, it can be a helpful guideline under challenging conditions. 

\noindent\textbf{LiDAR-camera Fusion.} As the recent examples, the fusion of LiDAR and camera features have shown successful demonstrations in various tasks such as 3D detection \cite{wang2021pointaugmenting, bai2022transfusion, liang2022bevfusion, wang2019pseudo, liang2018deep, xu2018pointfusion, yang2022deepinteraction}, map construction \cite{hendy2020fishing, hu2021fiery, saha2022translating, li2022bevformer, li2022hdmapnet, man2023bev, borse2023x, wang2023lidar2map, zou2023unim}, and both tasks \cite{liu2023bevfusion, vora2020pointpainting, yin2021multimodal}. Among them, BEVFusions \cite{li2022hdmapnet, liang2022bevfusion} introduce a shared BEV space for efficient fusions. Despite their promising perceptions, their BEV estimations in camera branches lack methods to share obtained sensing of each modality in the early stages. Although the models learn complementary fusion of BEV features, the spatially unsynchronized geometry between the sensor branches sometimes limits the range of model perceptions because the unlinked geometry causes worse interaction between modalities. In this work, we dig into a methodology for efficient sharing of modality sensings and a collaborative BEV fusion to effectively leverage geometry-synchronized sensors.

\section{Problem Formulation} \label{sec:formulation}
We notate $I$ and $P$ as camera and LiDAR modalities. The descriptions are under the assumption that image features ($\mathbf{Z}_I \in \mathbb{R}^{N\times H\times W\times C_I}$) and a LiDAR BEV feature ($\mathbf{Z}^{BEV}_P \in \mathbb{R}^{H'\times W'\times C_P}$) are prepared by Liu \textit{et al}'s method. \cite{liu2023bevfusion}

\vspace{4pt}\noindent\textbf{Conventional Method.} Most existing approaches \cite{liang2022bevfusion, liu2023bevfusion, borse2023x, man2023bev} obtain a unified BEV feature $\mathbf{Z}\in \mathbb{R}^{H'\times W'\times C_B}$ as
\begin{gather} \label{eq:legacy}
    \mathbf{Z} = G_\theta(\mathbf{C}, \mathbf{D}_I, \mathbf{Z}^{BEV}_P), \;\; \textrm{where}\;\; [\mathbf{C}; \mathbf{D}_I] = E_\omega(\mathbf{Z}_I),
\end{gather}
$G_\theta$ is a fusing model that consists of BEV pooling \cite{liu2023bevfusion} and deep nets such as CNN, transformer, and feature pyramid network (FPN) \cite{lin2017feature} with parameter $\theta$. $\mathbf{C}$ and $\mathbf{D}_I$ respectively denote an image context and a depth distribution \cite{philion2020lift}. $E_\omega := \mathbb{R}^{C_I}\mapsto (\mathbb{R}^{C_I}, \mathbb{R}^{C_D})$ is a single CNN with parameter set $\{\omega_1, \omega_2\}\subset\omega \subset \theta$. $\omega_1$ and $\omega_2$ are for $\mathbf{C}$ and $\mathbf{D}_I$, respectively.

\vspace{4pt}\noindent\textbf{BroadBEV.} As shown in \cref{eq:legacy}, the lack of hint for estimation of depth distribution sometimes becomes a bottleneck for a camera branch. As a result, the design shows weakness at night and rainy scenes because of harmful elements for depth estimation like limited camera eyesight and raindrops with unnatural textures. To this end, our strategy is to reduce the domain gap between a camera and a LiDAR BEV space. Specifically, our Point-scattering provides LiDAR BEV distribution $\mathbf{D}_P\in \mathbb{R}^{H'\times W'}$ to the camera depth distribution to guarantee a BEV space with synchronized geometry. The design addresses the concern that the scattering of LiDAR points to perspective camera planes causes poor compatibility with CNNs as shown by Wang \textit{et al.} \cite{wang2019pseudo}. After BroadBEV prepares camera and LiDAR BEV features, our ColFusion ($\mathbf{G}_\theta$) fuses them with shared attention weights for a robust BEV representation in real environments. The overall formulation of our BroadBEV is represented as
\begin{gather} \label{eq:BroadBEV}
    \mathbf{Z} = G_\theta(\mathbf{C}, \mathbf{D}_I, \mathbf{D}^{BEV}_P, \mathbf{Z}^{BEV}_P), \\
    \textrm{where} \;\; [\mathbf{C}; \mathbf{D}_I]=E_\omega(\mathbf{Z}_I), \;\; \mathbf{D}^{BEV}_P = \sigma(h_\nu(\mathbf{Z}^{BEV}_P)), \nonumber
\end{gather}
$h_{\nu}$ is a single CNN layer with a parameter set $\nu$, $\sigma$ is a sigmoid operation. $\nu \subset \theta$ is a set of parameters for the estimation of LiDAR BEV distribution.

\begin{figure}[t]
    \centering
    \begin{subfigure}[b]{0.77\linewidth}
        \centering
        \includegraphics[width=0.82\textwidth]{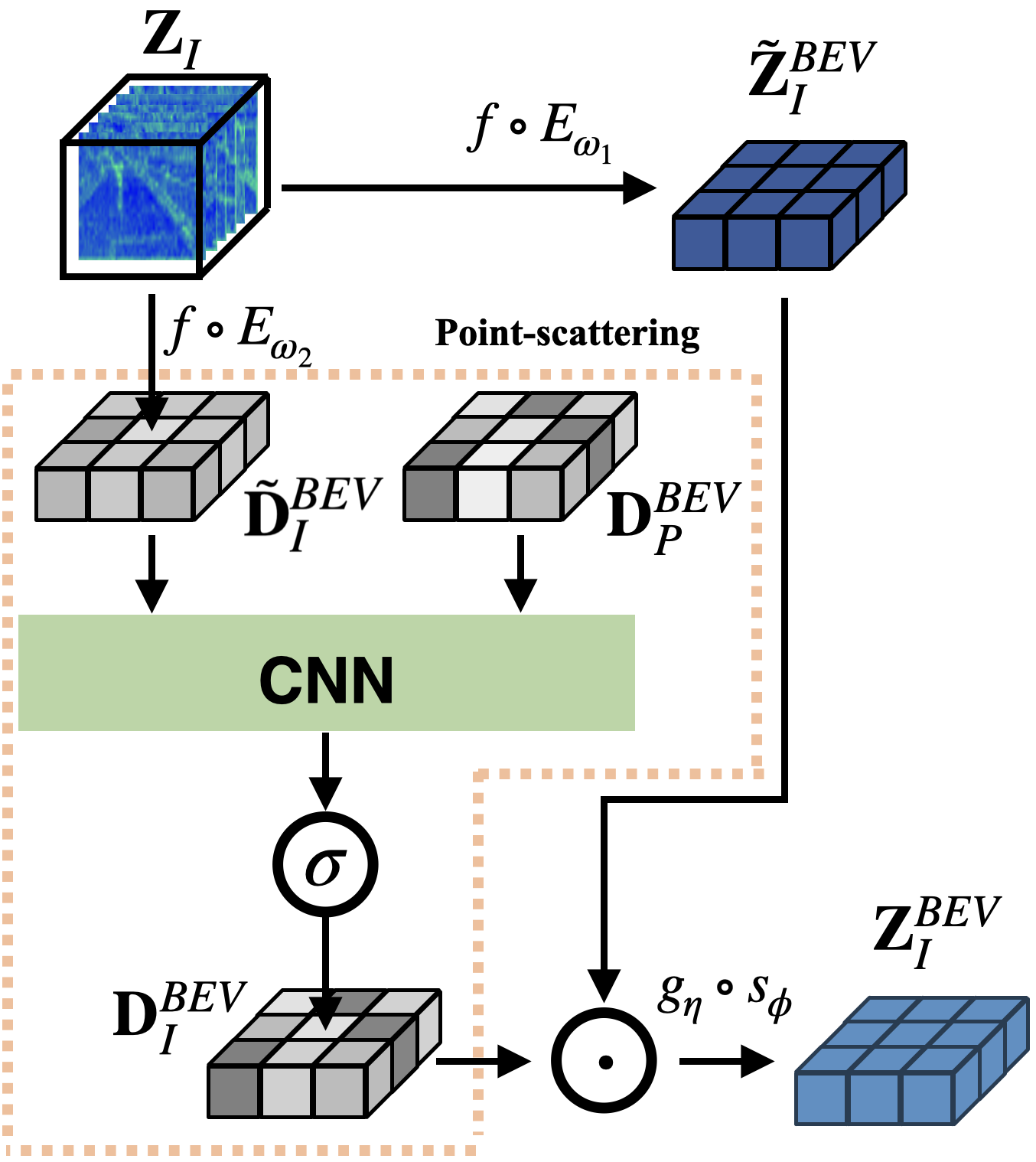}
    \end{subfigure}
    \vspace{5pt}
    \caption{\textbf{Our Point-scattering and View Transform.}}
    \label{fig:depth}
    \vspace{-16pt}
\end{figure}

\section{Method} \label{sec:method}
This section provides details of our formulation in \cref{eq:BroadBEV}. We first describe our novel view transform that guarantees depth synchronization between cross-modality. After that, we present descriptions for a collaborative BEV fusion (or ColFusion). The overall scheme of BroadBEV is illustrated in \cref{fig:flowchart}. As shown there, our BroadBEV takes in N cameras' image features ($\mathbf{Z}_I$), and a LiDAR BEV feature ($\mathbf{Z}^{BEV}_P$). After view transform and fusion, a semantic map is constructed.

\subsection{View Transform}
Compared to a camera, LiDAR shows consistent sensing in challenging conditions like signals damaged by raindrops, and limited eyesight by low-light illuminations. Our view transform aims to contain the advantage in camera BEV features. To implement the idea, we lift and splat $\mathbf{D}_I$ and interact it with LiDAR BEV distribution $\mathbf{D}^{BEV}_P$ in BEV space. The overall pipeline is provided in \cref{fig:depth}.

\begin{figure}[t]
    \centering
    \begin{subfigure}[b]{0.87\linewidth}
        \centering
        \includegraphics[width=\textwidth]{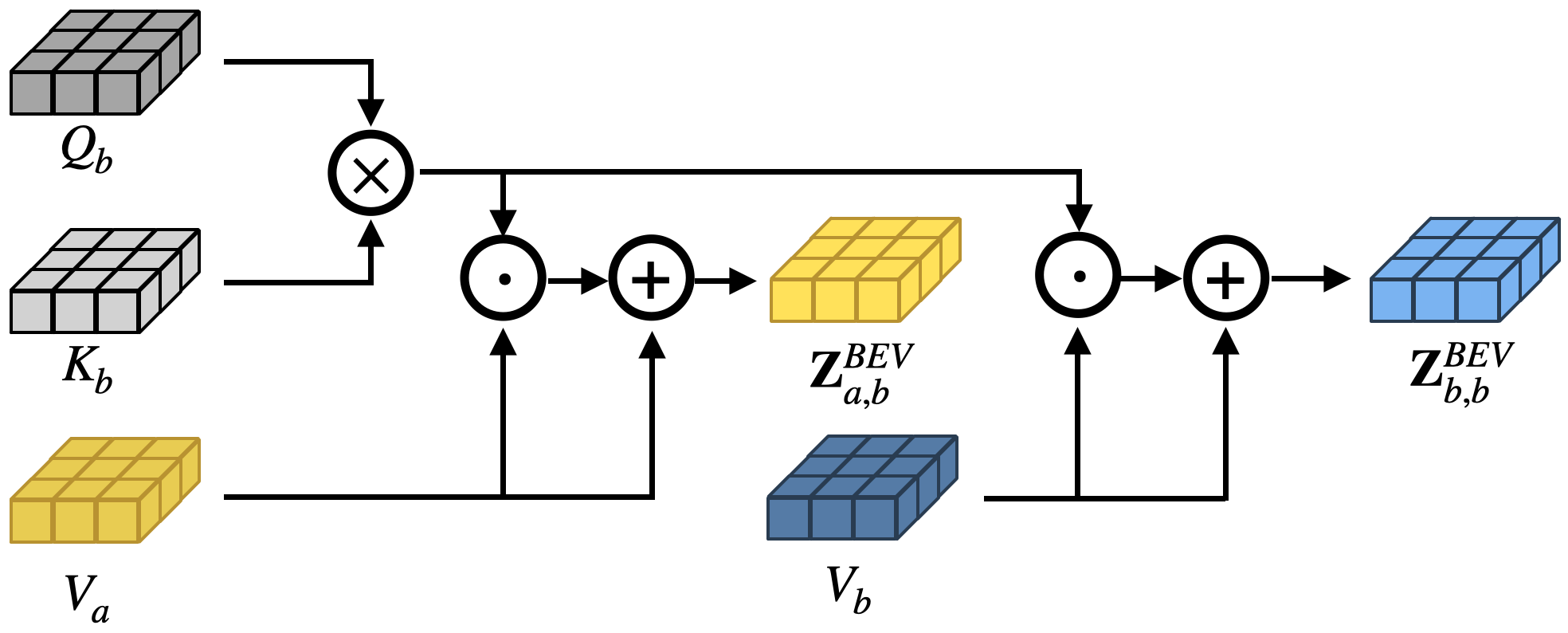}
        \caption{Cross Modality Attention.}
        \vspace{10pt}
    \end{subfigure}
    \begin{subfigure}[b]{0.66\linewidth}
        \centering
        \includegraphics[width=\textwidth]{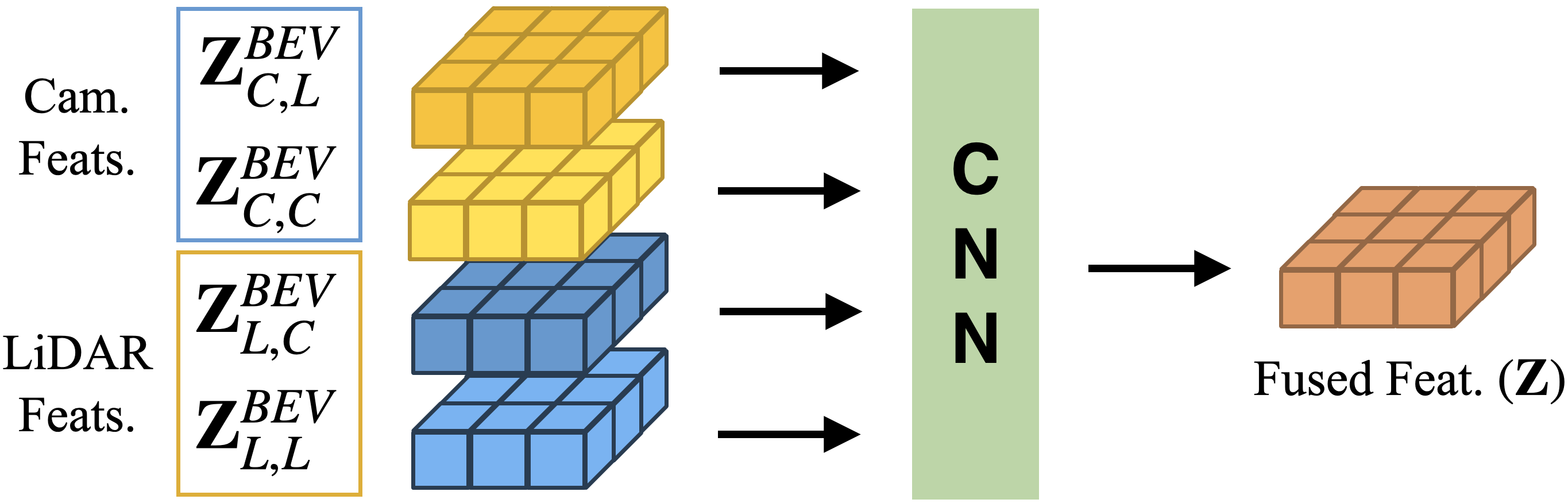}
        \caption{Extraction of a Unified BEV Feature.}
    \end{subfigure}
    \caption{\textbf{Our Collaborative BEV Fusion (ColFusion).}}
    \label{fig:fusion}
    \vspace{-18pt}
\end{figure}

\begin{table*}[t]
    \scriptsize
    \centering
    \caption{\textbf{Quantitative Comparisons on BEV Map Segmentation.}}
    \vspace{-5pt}
    \setlength{\tabcolsep}{1.5pt}
    \renewcommand{\arraystretch}{1.2}
    \begin{tabular}{
        c|>{\centering\arraybackslash}p{0.32\columnwidth}
        |>{\centering\arraybackslash}p{0.12\columnwidth}
        |>{\centering\arraybackslash}p{0.17\columnwidth}
        >{\centering\arraybackslash}p{0.17\columnwidth}
        >{\centering\arraybackslash}p{0.17\columnwidth}
        >{\centering\arraybackslash}p{0.17\columnwidth}
        >{\centering\arraybackslash}p{0.17\columnwidth}
        >{\centering\arraybackslash}p{0.17\columnwidth}
        |>{\centering\arraybackslash}p{0.17\columnwidth}
    }
        \Xhline{2\arrayrulewidth}
        Method & Backbone & Modality & Drivable & Ped. Cross. & Walkway & Stop Line & Carpark & Divider & mIoU \\
        \hline
        OFT \cite{roddick2018orthographic} & ResNet-18 & C & 74.0 & 35.3 & 45.9 & 27.5 & 35.9 & 33.9 & 42.1 \\
        LSS \cite{philion2020lift} & ResNet-18 & C & 75.4 & 38.8 & 46.3 & 30.3 & 39.1 & 36.5 & 44.4 \\
        CVT \cite{zhou2022cross} & EfficientNet-B4 & C & 74.3 & 36.8 & 39.9 & 25.8 & 35.0 & 29.4 & 40.2 \\
        M$^2$BEV \cite{xie2022m} & ResNeXt-101 & C & 77.2 & - & - & - & - & 40.5 & - \\
        BEVFusion \cite{liu2023bevfusion} & Swin-T & C & 81.7 & 54.8 & 58.4 & 47.4 & 50.7 & 46.4 & 56.6 \\
        \hline
        PointPillars \cite{lang2019pointpillars} & VoxelNet & L & 72.0 & 43.1 & 53.1 & 29.7 & 27.7 & 37.5 & 43.8 \\
        CenterPoint \cite{yin2021center} & VoxelNet & L & 75.6 & 48.4 & 57.5 & 36.5 & 31.7 & 41.9 & 48.6 \\
        \hline
        PointPainting \cite{vora2020pointpainting} & ResNet-101, PointPillars & C + L & 75.9 & 48.5 & 57.1 & 36.9 & 34.5 & 41.9 & 49.1 \\
        MVP \cite{yin2021multimodal} & ResNet-101, VoxelNet & C + L & 76.1 & 48.7 & 57.0 & 36.9 & 33.0 & 42.2 & 49.0 \\
        BEVFusion \cite{liu2023bevfusion} & Swin-T, VoxelNet & C + L & 85.5 & 60.5 & 67.6 & 52.0 & 57.0 & 53.7 & 62.7 \\
        X-Align \cite{borse2023x} & Swin-T, VoxelNet & C + L & 86.8 & 65.2 & 70.0 & 58.3 & 57.1 & 58.2 & 65.7 \\
        UniM$^2$AE \cite{zou2023unim} & Swin-T, SST & C + L & 88.7 & 67.4 & 72.9 & 59.0 & 59.0 & 59.7 & 67.8 \\
        \rowcolor[gray]{0.8} BroadBEV (Ours) & Swin-T, VoxelNet & C + L & \textbf{90.1} & \textbf{69.4} & \textbf{75.9} & \textbf{60.2} & \textbf{64.2} & \textbf{60.8} & \textbf{70.1} \\
        \Xhline{2\arrayrulewidth}
    \end{tabular}
    \label{tab:Quan1}
\end{table*}
\begin{table*}[t]
    \scriptsize
    \centering
    \caption{\textbf{Quantitative Comparisons on HD Map Construction.}}
    \vspace{-5pt}
    \setlength{\tabcolsep}{1.5pt}
    \renewcommand{\arraystretch}{1.2}
    \begin{tabular}{
        c|>{\centering\arraybackslash}p{0.45\columnwidth}
        |>{\centering\arraybackslash}p{0.25\columnwidth}
        |>{\centering\arraybackslash}p{0.24\columnwidth}
        >{\centering\arraybackslash}p{0.24\columnwidth}
        >{\centering\arraybackslash}p{0.24\columnwidth}
        |>{\centering\arraybackslash}p{0.22\columnwidth}
    }
        \Xhline{2\arrayrulewidth}
        Method & Backbone & Modality & Divider & Ped. Cross. & Boundary & mIoU \\
        \hline
        VPN \cite{pan2020cross} & EfficientNet-B0 & C & 36.5 & 15.8 & 35.6 & 29.3 \\
        Lift-Splat \cite{philion2020lift} & EfficientNet-B0 & C & 38.3 & 14.9 & 39.3 & 30.8 \\
        BEVSegFormer \cite{peng2023bevsegformer} & ResNet-101 & C & 51.1 & 32.6 & 50.0 & 44.6 \\
        BEVFormer \cite{li2022bevformer} & ResNet-50 & C & 53.0 & 36.6 & 54.1 & 47.9 \\
        BEVerse \cite{zhang2022beverse} & Swin-T & C & 56.1 & 44.9 & 58.7 & 53.2 \\
        UniFusion \cite{qin2022uniformer} & Swin-T & C & 58.6 & 43.3 & 59.0 & 53.6 \\
        HDMapNet \cite{li2022hdmapnet} & EfficientNet-B0 & C & 40.6 & 18.7 & 39.5 & 32.9 \\
        \hline
        HDMapNet \cite{li2022hdmapnet} & PointPillars & L & 26.7 & 17.3 & 44.6 & 29.5 \\
        LiDAR2Map \cite{wang2023lidar2map} & PointPillars & L & 60.4 & 45.5 & 66.4 & 57.4 \\
        \hline
        HDMapNet \cite{li2022hdmapnet} & EfficientNet-B0, PointPillars & C + L & 46.1 & 31.4 & 56.0 & 44.5 \\
        LiDAR2Map \cite{wang2023lidar2map} & Swin-T, PointPillars & C + L & 60.8 & 47.2 & 66.3 & 58.1 \\
        \rowcolor[gray]{0.8} BroadBEV (Ours) & Swin-T, VoxelNet & C + L & \textbf{68.8} & \textbf{51.2} & \textbf{71.9} & \textbf{64.0} \\
        \Xhline{2\arrayrulewidth}
    \end{tabular}
    \label{tab:Quan2}
    \vspace{-8pt}
\end{table*}
\begin{table}[t]
    \centering
    \scriptsize
    \vspace{-4pt}
    \caption{\textbf{Comparison of Map Segmentation to state-of-the-arts under Rainy or Night Condition.}}
    \begin{subtable}[t]{\linewidth}
        \centering
        \vspace{-5pt}
        \begin{tabular}{
            c
            |>{\centering\arraybackslash}p{0.20\linewidth}
            >{\centering\arraybackslash}p{0.20\linewidth}
            |>{\centering\arraybackslash}p{0.20\linewidth}
        }
            \Xhline{2\arrayrulewidth}
            \multirow{2}{*}{Method} & \multirow{2}{*}{Rainy} & \multirow{2}{*}{Night} & \multirow{2}{*}{nuScenes} \\
             & & & \\
            \hline
            BEVFusion & 55.9 & 43.6 & 62.7 \\
            X-Align & 57.8 & 46.1 & 65.7 \\
            \rowcolor[gray]{0.8} BroadBEV (Ours) & \textbf{63.7} & \textbf{50.8} & \textbf{70.1} \\
            \Xhline{2\arrayrulewidth}
        \end{tabular}
    \end{subtable}
    \label{tab:Quan3}
    \vspace{-14pt}
\end{table}

\vspace{4pt}\noindent\textbf{Lift-Splat \cite{philion2020lift}.} At first, we define two variables: a camera BEV feature without depth distribution ($\Tilde{\mathbf{Z}}_I^{BEV}$), and a depth distribution without image features ($\Tilde{\mathbf{D}}_I^{BEV} \in \mathbb{R}^1$). They are prepared as
\begin{gather}
    \Tilde{\mathbf{Z}}_I^{BEV} = f(\mathbf{C}, \mathds{1}), \quad
    \Tilde{\mathbf{D}}_I^{BEV} = f(\mathds{1}, \mathbf{D}_I), \label{eq:pool}
\end{gather}
where $\mathds{1}$ is a $H\times W$ tensor of ones. $f$ denotes a BEV pooling, i.e., $f(\mathbf{Z}_I, \mathbf{D}_I)$ is the view transformation of Liu \textit{et al.} \cite{liu2023bevfusion}.

\vspace{4pt}\noindent\textbf{Point-scattering.} After preparing variables, BroadBEV scatters $\mathbf{D}_P$ to a depth distribution $\Tilde{\mathbf{D}}_I^{BEV}$ as
\begin{gather}
    \mathbf{D}_I^{BEV} = \sigma(h_\psi(\Tilde{\mathbf{D}}_I^{BEV}, \mathbf{D}^{BEV}_P)),
\end{gather}
where $h_\psi := (\mathbb{R}^1, \mathbb{R}^1) \mapsto \mathbb{R}^{C_I}$ is a CNN layer, $\mathbf{D}^{BEV}_I$ is a refined camera depth distribution.

\vspace{4pt}\noindent\textbf{Feature Calibration.} As a LiDAR BEV represents sparse shapes while a camera BEV contains dense image features, there is an inevitable domain gap between modalities. To address the problem, we complete the camera BEV feature $\mathbf{Z}_I^{BEV}$ with a self-calibrated convolution ($s_\phi$) \cite{liu2020improving} as
\begin{gather}
    \mathbf{Z}_I^{BEV} = g_\eta(s_\phi(\Tilde{\mathbf{Z}}_I^{BEV} \cdot \mathbf{D}_I^{BEV})),
\end{gather}
where $g_\eta$ is an FPN with the same hyperparameter configurations as \cite{liu2023bevfusion}.

\subsection{Collaborative Fusion}
After the preparation of BEV features ($\mathbf{Z}^{BEV}_I, \mathbf{Z}^{BEV}_P$), our devised collaborative fusion (or ColFusion) ensembles them. Specifically, ColFusion computes the self-attention weights of its modality branches and shares them with each other to effectively leverage the synchronized sensors. Then the model fuses the BEV features with deep nets. ColFusion's pipelines are illustrated in \cref{fig:fusion}.


\vspace{4pt}\noindent\textbf{Attention of Cross Modality.} When given two inputs of modality $a, b \in \{I, P\}$, we obtain a refined BEV feature $\mathbf{Z}^{BEV}_{b, a}$ from an attention weight ($\mathbf{A}_a$) and a BEV feature $\mathbf{Z}_b^{BEV}$. $\mathbf{A}_a$ is computed from a query ($Q_a$), and a key ($K_a$) \cite{vaswani2017attention} of $\mathbf{Z}_a^{BEV}$, and a value ($V_b$) is computed from $\mathbf{Z}_b^{BEV}$. The equation can be represented as
\begin{gather} \label{eq:bevfeat}
    \mathbf{A}_a = \textrm{softmax}(\frac{Q_a\cdot K_a^{T}}{\sqrt{d_a}}), \quad \mathbf{Z}_{b,\;a}^{BEV} = \mathbf{A}_a \cdot V_b + \mathbf{Z}^{BEV}_b,
\end{gather}
where $d_k$ denotes a scale factor \cite{vaswani2017attention}. In other words, we compute four BEV features ($\mathbf{Z}^{BEV}_{I,\; I}, \mathbf{Z}^{BEV}_{I,\; P}, \mathbf{Z}^{BEV}_{P,\; I}, \mathbf{Z}^{BEV}_{P,\; P}$) with shared self-attention weights and the prepared BEV features.

\vspace{4pt}\noindent\textbf{BEV Representation.} We unify the prepared 4 BEV features as a BEV representation $\mathbf{Z}$ as
\begin{gather}
    \mathbf{Z} = g_\varphi(\sum_{a,\; b} \mathbf{W}_{a,\;b}\cdot \mathbf{Z}_{a,\;b}^{BEV}),
\end{gather}
where $\mathbf{W}_{a,\;b}$ is a partial weight of CNN parameters. $g_\varphi$ denotes BEV-FPD \cite{wang2023lidar2map} which is a kind of FPN to encode the unified BEV feature. After BroadBEV obtains $\mathbf{Z}$, it decodes the feature with a task-specific head.

\section{Experimental Results} \label{sec:exp}
Following the previous works, we evaluate BroadBEV on map segmentation and HD map construction in \cref{tab:Quan1} and \cref{tab:Quan2}, respectively. we use mean Intersection over Union (mIoU) as our evaluation metric in all the experiments including our ablation study and visualizations.

\subsection{Configurations}
\noindent\textbf{Dataset.} We use nuScenes \cite{caesar2020nuscenes}, a large-scale dataset that contains 700, 150, and 150 scenes for training, validation, and testing. The collected samples include various data modalities such as perspective images from 6 cameras, 3D points from 1 LiDAR, and points with vectors from 5 Radars. We use camera images resized to $256\times704$ resolution, and LiDAR points voxelized to 0.1m grid resolution in our LiDAR BEV branch.

\noindent\textbf{Implementation Details.} We use common configurations on both BEV map segmentation and HD map construction tasks. We apply the same image and LiDAR data augmentation as the baseline work \cite{liu2023bevfusion}. We use AdamW \cite{loshchilov2017decoupled} with a weight decay of $10^{-2}$. We train BroadBEV during 20 epochs on 4 NVIDIA A100 GPUs. We use 3 heads for multi-head attention of ColFusion. We employ Swin-T \cite{liu2021swin} (or VoxelNet \cite{zhou2018voxelnet}) as a backbone for the camera (or LiDAR) branch. Our implementations are built on top of mmdetections \cite{chen2019mmdetection}, \cite{mmdet3d2020}.

\begin{table}[t]
    \centering
    \scriptsize
    \caption{\textbf{Ablation Study on Method Specifications.}}
    \begin{subtable}[t]{\linewidth}
        \centering
        \vspace{-5pt}
        \begin{tabular}{
            c
            |>{\centering\arraybackslash}p{0.08\linewidth}
            >{\centering\arraybackslash}p{0.12\linewidth}
            |>{\centering\arraybackslash}p{0.07\linewidth}
            >{\centering\arraybackslash}p{0.07\linewidth}
            >{\centering\arraybackslash}p{0.09\linewidth}
            |>{\centering\arraybackslash}p{0.09\linewidth}
        }
            \Xhline{2\arrayrulewidth}
            \multirow{2}{*}{Baseline} & \multirow{2}{*}{w/ Scat.} & \multirow{2}{*}{w/ Fusion} & \multirow{2}{*}{Rainy} & \multirow{2}{*}{Night} & \multirow{2}{*}{nuScenes} & Latency \\
             & & & & & & (ms) \\
            \hline
            \ding{51} & & & 56.3 & 43.8 & 62.6 & 83 \\
            \ding{51} & \ding{51} &  & 57.4 & 45.7 & 63.9 & 88 \\
            \ding{51} & & \ding{51} & 62.6 & 50.1 & 69.1 & 152 \\
            \rowcolor[gray]{0.8} \ding{51} & \ding{51} & \ding{51} & \textbf{63.7} & \textbf{50.8} & \textbf{70.1} & 158 \\
            \Xhline{2\arrayrulewidth}
        \end{tabular}
    \end{subtable}
    \label{tab:ablation}
    \vspace{-14pt}
\end{table}


\begin{figure*}[t]
    \centering
    \begin{subfigure}[b]{\linewidth}
        \centering
        \includegraphics[width=0.87\linewidth]{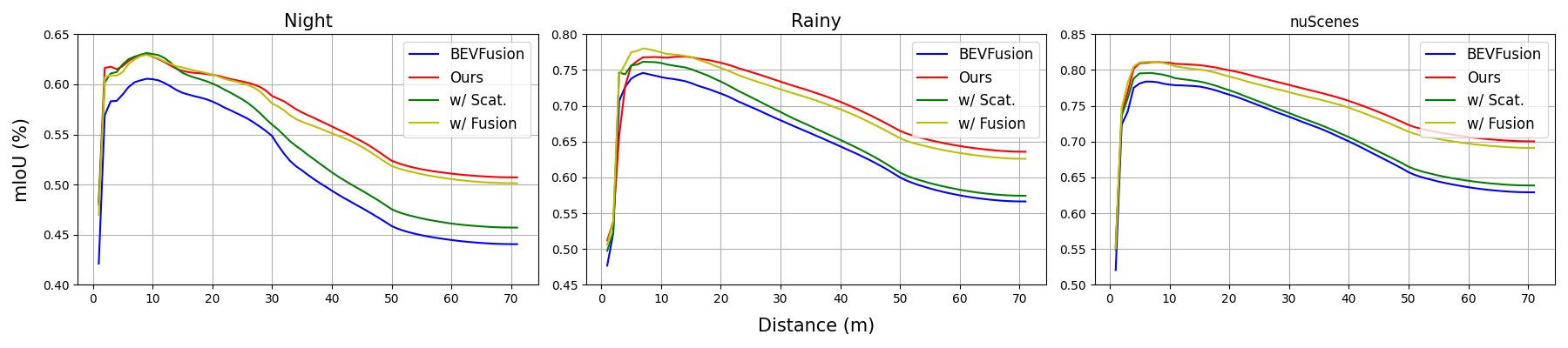}
        \vspace{-5pt}
    \end{subfigure}
    \caption{\textbf{Averaged Performances for Varying Distances from Ego Vehicle.}}
    \label{fig:distance_graph}
    \vspace{-14pt}
\end{figure*}

\begin{figure}[t]
    \centering
    \includegraphics[width=0.8\linewidth]{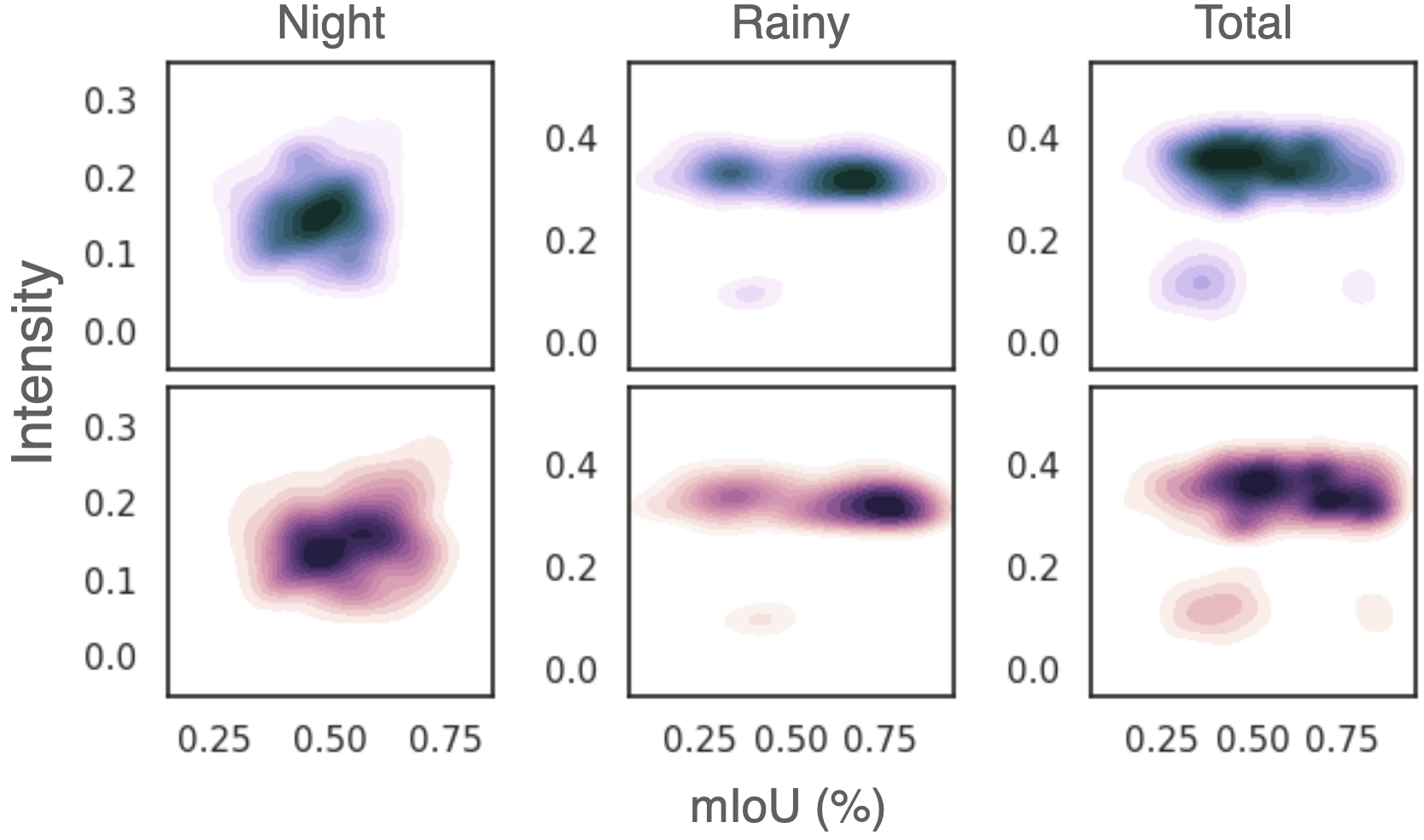}
    \vspace{-4pt}
    \caption{\textbf{Performance Distribution under Varying Illuminations.} Blue and pink plots correspond to BEVFusion and Ours, respectively. Given average image intensities, ours achieves improved map constructions showing right-shifted distribution than BEVFusion.}
    \label{fig:illumination}
    \vspace{-20pt}
\end{figure}

\subsection{BEV Map Segmentation}
\noindent\textbf{Settings.} We follow the evaluation protocol of \cite{liu2023bevfusion} under 100m$\times$100m with 0.5m egocentric BEV grid resolution. The employed metric measures mIoU for each class and then chooses the highest value over varying thresholds as the semantic map classes of urban environments may be semantically overlapped like car-parking and drivable areas.

\noindent\textbf{Overall Comparisons.} We report comparisons of BroadBEV to the state-of-the-arts in \cref{tab:Quan1}, and \cref{tab:Quan3}. The former focuses on per-class comparisons of method performances, and the latter provides the effectiveness of approaches under rainy, night conditions. In \cref{tab:Quan1}, we investigate the employed backbones of all methods including ResNet \cite{he2016deep}, ResNext \cite{xie2017aggregated}, EfficientNet \cite{tan2019efficientnet}, Swin Transformer \cite{liu2021swin}, VoxelNet \cite{zhou2018voxelnet}, PointPillars \cite{lang2019pointpillars}, and SST \cite{fan2022embracing}. As shown in the table, our method outperforms existing approaches without bells and whistles in every class. Especially, considering that map segmentation is more dependent on dense features than sparse LiDAR features, the remarkable achievements in the classes imply that BroadBEV provides improved fusion with enhanced camera BEV features. In \cref{tab:Quan3}, our methods show favorable performance gains in rainy and night conditions. The results support that our methods contribute to model robustness and performance consistency.

\begin{figure*}[t]
    \centering
    \begin{subfigure}[b]{\linewidth}
        \centering
        \includegraphics[width=0.85\linewidth]{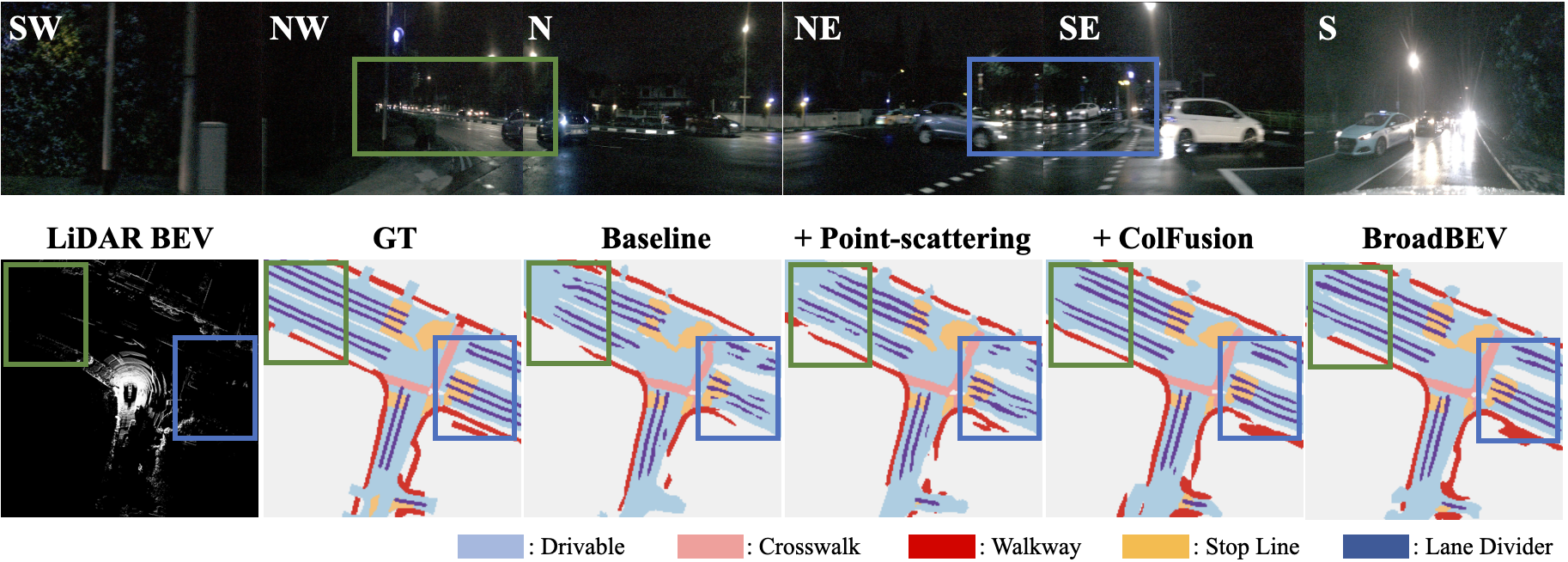}
    \end{subfigure}
    \caption{\textbf{Ablation Study on Broadness.} The upper row displays RGB images. The lower row contains LiDAR BEV and map construction of our variated models. ``\textbf{Baseline}", ``\textbf{+\;Point-scattering}", and ``\textbf{+\;ColFusion}" respectively correspond to ``Baseline", ``w/ Scat.", and ``w/ Fusion" models in \cref{tab:ablation}. Green ROIs show Point-scattering's contribution to broad-sighted perception. Blue ROIs demonstrate ColFusion's enhancement of perception performance in distant regions.}
    \label{fig:abl_broad}
    \vspace{-10pt}
\end{figure*}

\noindent\textbf{Ablation Study.} We explore the effectiveness of Point-scattering and ColFusion to check their importance. In \cref{tab:ablation}, ``w/ Fusion" and ``w/ Scat." mean ColFusion and Point-scattering respectively. We denote `\ding{51}' if a method is activated. As in the table, our fusion largely boosts perception performance. In \cref{fig:abl_broad}, we evaluate the models of \cref{tab:ablation} to intuitively check module-by-module contributions. As shown in the camera images, the driving environment is under humid night which causes a performance drop. As demonstrated in the green ROIs, Point-scattering improves the perception broadness. Due to the large sparsity of LiDAR points in distant regions, the perception in the areas is dominantly inferred by camera branches. Despite this fact, BroadBEV shows favorable map segmentation owing to the Point-scattering's successful boosting of the camera branch weights in the training stage. Because the results show our accurate locating of scene contexts without any Lidar priors in the green boxes, they further validate our learning of camera depth refinement for completion of entire spatial coverage. In addition, ColFusion enhances the perception of distant regions as in the blue ROIs. Thanks
to its collaborative modality fusion, the method further boosts BroadBEV's broadness.

\noindent\textbf{Computation Time.} In \cref{tab:ablation}, we compare the computing cost of BroadBEV to our variations under single A100 GPU. Note that ``baseline" has the same design as BEVFusion \cite{liu2023bevfusion}. As in the table, ColFusion imposes dominant costs (about 70ms of 158ms) because of the heavy attention mechanism. We expect that a replacement of the attention with a light operation as in PoolFormer \cite{yu2022metaformer} will address the problem.

\noindent\textbf{Distance Dependency.} In \cref{fig:distance_graph}, we explore a detailed comparison of our perception broadness to BEVFusion and our variated models used in our ablation study. The horizontal xx-axis and vertical yy-axis respectively mean range [1,x][1, x] meter distances from an egocentric vehicle and the corresponding mIoU. In this experiment, we see that our method regularizes the negative correlation between map segmentation and distance. Especially, considering that LiDAR mostly preserves its sensing at night, the performance gain in ``Night" plot validates the contribution of LiDAR Point-scattering for robust perception under low-light illumination. Furthermore, our performance gains in distant regions imply that BroadBEV provides a broad-sighted BEV representation.

\noindent\textbf{Illumination Dependency.} In \cref{fig:illumination}, we explore the relationship between the fusion approaches and illumination. We visualize distributions of scattered mIoUs for all validation samples. The densely (or sparsely) distributed regions are colored with darker (or more light) tones. To evaluate the illumination of a sample, we transform RGB to YUV and average Y channel. The horizontal $x$-axis and vertical $y$-axis respectively mean mIoU and averaged Y values (or intensity). In the plots, our Point-scattering and ColFusion provide the improved performance distributions that are located on the higher mIoU regions.  Especially, the ``Night" plot shows a promising demonstration implying that our fusion resolves the limited generalization of a camera at night.

\subsection{HD Map Construction}
HD map is a pre-constructed information hub that consists of descriptions of huge 3D volumes like buildings with many indoor, and urban environments. It includes 3D bounding boxes, semantic labels, vectorized road lanes, etc. Our experiment on this task follows the previous works \cite{li2022hdmapnet, wang2023lidar2map} that aim for auto-labeling or maintenance of semantic road labels in HD maps by constructing locally constructed maps.

\noindent\textbf{Settings.} Because the HD map contains densely distributed 3D points, its label prediction is deployed on a dense 30m$\times$60m volume with 0.15m BEV grid resolution. Following the metric of the previous works \cite{li2022hdmapnet}, \cite{wang2023lidar2map}, we use mIoU without the thresholding of the map segmentation task.

\noindent\textbf{Overall Comparisons.} The comparisons of BroadBEV to the existing HD map constructors are provided in \cref{tab:Quan2}. As indicated in the table, our model shows state-of-the-art performance with $5.9\%$ (Divider: $8.0\% \uparrow$, Ped.: $4.0\% \uparrow$, Bnd.: $5.6\% \uparrow$) mIoU gain compared to the LiDAR2Map. Because HD map requires high-frequency detailed local descriptions for the surrounding environment, our performance gain implies BroadBEV's superior extraction of local features.

\begin{figure}[t]
    \centering
    \begin{subfigure}[b]{\linewidth}
        \centering
        \includegraphics[width=0.70\linewidth]{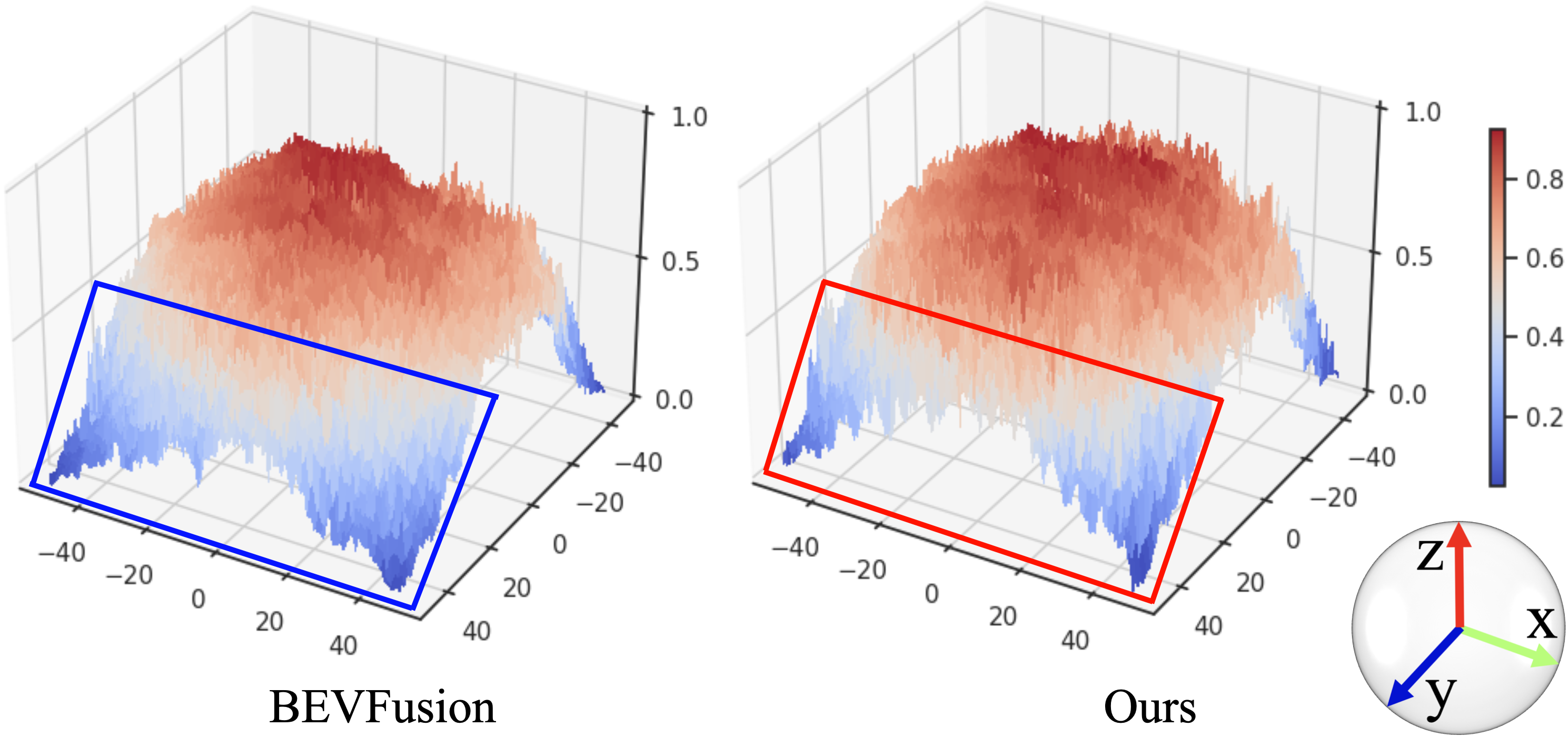}
        \subcaption{Observation on mIoU Distribution at BEV Edges.}
        \label{fig:distance_viz} \vspace{4pt}
    \end{subfigure}
    \begin{subfigure}[b]{\linewidth}
        \centering
        \includegraphics[width=0.75\linewidth]{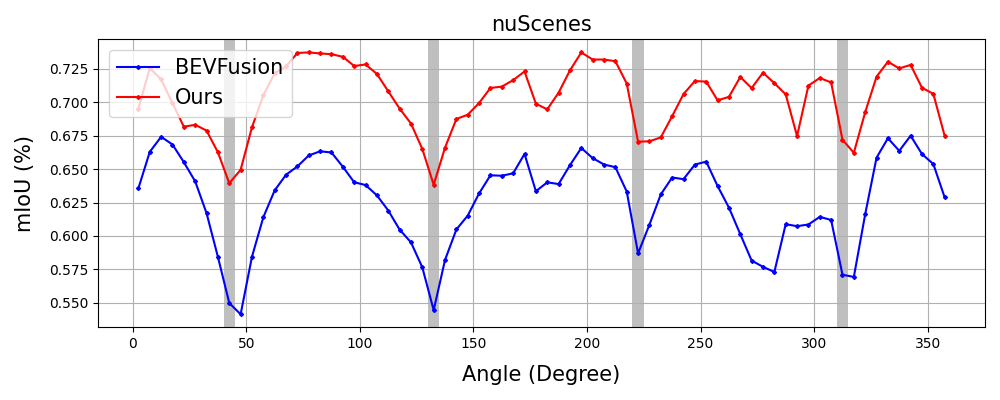}
        \vspace{-4pt}
        \subcaption{Performance Undershoot caused by BEV Vertices.}
        \label{fig:angle_viz}
    \end{subfigure}
    \vspace{-14pt}
    \caption{\textbf{Visualization of BEV Edges and Vertices.} (a) x and y axes indicate distances from an egocentric vehicle frame. z-axis is the mIoU averaged over all samples. (b) Gray shadings are the angles containing the BEV vertices.}
    \vspace{-17pt}
\end{figure}

\section{Discussion} \label{sec:discussion}
In this section, we analyze and discuss a limitation of BEV perception. The results are investigated using all the samples of the nuScenes validation set.

\vspace{4pt}\noindent\textbf{Performance Undershoot in BEV Vertices.} As in \cref{fig:distance_viz}, BEV features provide relatively poor representations on the vertices. To explore how the limitation negatively affects map segmentation, we investigate performance distribution for varying scanning angles in \cref{fig:angle_viz}. From an egocentric vehicle frame, we divide a map into 72 segments with 5$^\circ$ field-of-view (FOV), i.e., the $y$ value of each point denotes an averaged mIoU in an FOV segment. As depicted in the figure, the regions containing BEV vertices (gray regions) show significant performance drops. In fact, for objects at very far distances, LiDAR often fails to detect them. Furthermore, the mostly vanished shape of them in camera planes is one of the challenging limitations for learning BEV representation. Although our broadness addressed performance degradation, still the problem is a big hurdle to cross.

\vspace{4pt}\noindent\textbf{Larger Camera Frustum} Extending camera frustums \cite{philion2020lift} relaxes the undershoot. However, the approach fails to resolve the limitation. In \cref{tab:frustum}, we explore two BroadBEV models under 60m (BroadBEV) and 70m (Variation) frustum depth configurations. As in the table, ``Variation'' shows a slightly better segmentation but could not resolve the problem. Future works addressing this issue will be interesting.

\begin{table}[t]
    \centering
    \small
    \caption{\textbf{Frustum Depth Range and Performance.}}
    \begin{subtable}[t]{\linewidth}
        \centering
        \vspace{-3pt}
        \begin{tabular}{
            c
            |>{\centering\arraybackslash}p{0.40\linewidth}
            |>{\centering\arraybackslash}p{0.25\linewidth}
        }
            \Xhline{2\arrayrulewidth}
            Method & Range & mIoU \\
            \hline
            Variation & (-70.0m, 70.0m) & 70.2 \\
            \rowcolor[gray]{0.8} BroadBEV & (-60.0m, 60.0m) & 70.1 \\
            \Xhline{2\arrayrulewidth}
        \end{tabular}
    \end{subtable}
    \label{tab:frustum}
    \vspace{-14pt}
\end{table}


\section{Conclusion}
Our proposed method BroadBEV provides a broad-sighted bird's eye view representation. The synchronized geometry between cross-modality contributes to the precise location of dense contexts in the proper position of a BEV space. In addition, our ColFusion ensembles LiDAR and camera BEV features collaboratively with a novel attention mechanism-based fusion. Compared to the recent model, X-Align, BroadBEV shows noticeable improvements in overall driving conditions including rainy, night (or low-light), and whole data scenarios with 5.9\%, 4.7\%, and 4.4\% mIoU gains, respectively. Furthermore, our model additionally validated its robustness on HD map construction, which requires high-frequency detailed features. In the task, we achieved 5.9\% mIoU gain compared to LiDAR2Map.




\bibliographystyle{ieee_fullname}
\bibliography{egbib}

\begin{thebibliography}{10}\itemsep=-1pt

\bibitem{bai2022transfusion}
Xuyang Bai, Zeyu Hu, Xinge Zhu, Qingqiu Huang, Yilun Chen, Hongbo Fu, and
  Chiew-Lan Tai.
\newblock Transfusion: Robust lidar-camera fusion for 3d object detection with
  transformers.
\newblock In {\em Proceedings of the IEEE/CVF conference on computer vision and
  pattern recognition}, pages 1090--1099, 2022.

\bibitem{borse2023x}
Shubhankar Borse, Marvin Klingner, Varun~Ravi Kumar, Hong Cai, Abdulaziz
  Almuzairee, Senthil Yogamani, and Fatih Porikli.
\newblock X-align: Cross-modal cross-view alignment for bird's-eye-view
  segmentation.
\newblock In {\em Proceedings of the IEEE/CVF Winter Conference on Applications
  of Computer Vision}, pages 3287--3297, 2023.

\bibitem{caesar2020nuscenes}
Holger Caesar, Varun Bankiti, Alex~H Lang, Sourabh Vora, Venice~Erin Liong,
  Qiang Xu, Anush Krishnan, Yu Pan, Giancarlo Baldan, and Oscar Beijbom.
\newblock nuscenes: A multimodal dataset for autonomous driving.
\newblock In {\em Proceedings of the IEEE/CVF conference on computer vision and
  pattern recognition}, pages 11621--11631, 2020.

\bibitem{carion2020end}
Nicolas Carion, Francisco Massa, Gabriel Synnaeve, Nicolas Usunier, Alexander
  Kirillov, and Sergey Zagoruyko.
\newblock End-to-end object detection with transformers.
\newblock In {\em European conference on computer vision}, pages 213--229.
  Springer, 2020.

\bibitem{chen2019mmdetection}
Kai Chen, Jiaqi Wang, Jiangmiao Pang, Yuhang Cao, Yu Xiong, Xiaoxiao Li,
  Shuyang Sun, Wansen Feng, Ziwei Liu, Jiarui Xu, et~al.
\newblock Mmdetection: Open mmlab detection toolbox and benchmark.
\newblock {\em arXiv preprint arXiv:1906.07155}, 2019.

\bibitem{mmdet3d2020}
MMDetection3D Contributors.
\newblock {MMDetection3D: OpenMMLab} next-generation platform for general {3D}
  object detection.
\newblock \url{https://github.com/open-mmlab/mmdetection3d}, 2020.

\bibitem{fan2022embracing}
Lue Fan, Ziqi Pang, Tianyuan Zhang, Yu-Xiong Wang, Hang Zhao, Feng Wang, Naiyan
  Wang, and Zhaoxiang Zhang.
\newblock Embracing single stride 3d object detector with sparse transformer.
\newblock In {\em Proceedings of the IEEE/CVF conference on computer vision and
  pattern recognition}, pages 8458--8468, 2022.

\bibitem{hartley2003multiple}
Richard Hartley and Andrew Zisserman.
\newblock {\em Multiple view geometry in computer vision}.
\newblock Cambridge university press, 2003.

\bibitem{he2016deep}
Kaiming He, Xiangyu Zhang, Shaoqing Ren, and Jian Sun.
\newblock Deep residual learning for image recognition.
\newblock In {\em Proceedings of the IEEE conference on computer vision and
  pattern recognition}, pages 770--778, 2016.

\bibitem{hendy2020fishing}
Noureldin Hendy, Cooper Sloan, Feng Tian, Pengfei Duan, Nick Charchut, Yuesong
  Xie, Chuang Wang, and James Philbin.
\newblock Fishing net: Future inference of semantic heatmaps in grids.
\newblock {\em arXiv preprint arXiv:2006.09917}, 2020.

\bibitem{hong2022cross}
Yu Hong, Hang Dai, and Yong Ding.
\newblock Cross-modality knowledge distillation network for monocular 3d object
  detection.
\newblock In {\em European Conference on Computer Vision}, pages 87--104.
  Springer, 2022.

\bibitem{hu2021fiery}
Anthony Hu, Zak Murez, Nikhil Mohan, Sof{\'\i}a Dudas, Jeffrey Hawke, Vijay
  Badrinarayanan, Roberto Cipolla, and Alex Kendall.
\newblock Fiery: Future instance prediction in bird's-eye view from surround
  monocular cameras.
\newblock In {\em Proceedings of the IEEE/CVF International Conference on
  Computer Vision}, pages 15273--15282, 2021.

\bibitem{huang2022bevdet4d}
Junjie Huang and Guan Huang.
\newblock Bevdet4d: Exploit temporal cues in multi-camera 3d object detection.
\newblock {\em arXiv preprint arXiv:2203.17054}, 2022.

\bibitem{huang2021bevdet}
Junjie Huang, Guan Huang, Zheng Zhu, Yun Ye, and Dalong Du.
\newblock Bevdet: High-performance multi-camera 3d object detection in
  bird-eye-view.
\newblock {\em arXiv preprint arXiv:2112.11790}, 2021.

\bibitem{kim2019deep}
Youngseok Kim and Dongsuk Kum.
\newblock Deep learning based vehicle position and orientation estimation via
  inverse perspective mapping image.
\newblock In {\em 2019 IEEE Intelligent Vehicles Symposium (IV)}, pages
  317--323. IEEE, 2019.

\bibitem{lang2019pointpillars}
Alex~H Lang, Sourabh Vora, Holger Caesar, Lubing Zhou, Jiong Yang, and Oscar
  Beijbom.
\newblock Pointpillars: Fast encoders for object detection from point clouds.
\newblock In {\em Proceedings of the IEEE/CVF conference on computer vision and
  pattern recognition}, pages 12697--12705, 2019.

\bibitem{li2022hdmapnet}
Qi Li, Yue Wang, Yilun Wang, and Hang Zhao.
\newblock Hdmapnet: An online hd map construction and evaluation framework.
\newblock In {\em 2022 International Conference on Robotics and Automation
  (ICRA)}, pages 4628--4634. IEEE, 2022.

\bibitem{li2023bevdepth}
Yinhao Li, Zheng Ge, Guanyi Yu, Jinrong Yang, Zengran Wang, Yukang Shi,
  Jianjian Sun, and Zeming Li.
\newblock Bevdepth: Acquisition of reliable depth for multi-view 3d object
  detection.
\newblock In {\em Proceedings of the AAAI Conference on Artificial
  Intelligence}, volume~37, pages 1477--1485, 2023.

\bibitem{li2022bevformer}
Zhiqi Li, Wenhai Wang, Hongyang Li, Enze Xie, Chonghao Sima, Tong Lu, Yu Qiao,
  and Jifeng Dai.
\newblock Bevformer: Learning bird’s-eye-view representation from
  multi-camera images via spatiotemporal transformers.
\newblock In {\em European conference on computer vision}, pages 1--18.
  Springer, 2022.

\bibitem{liang2018deep}
Ming Liang, Bin Yang, Shenlong Wang, and Raquel Urtasun.
\newblock Deep continuous fusion for multi-sensor 3d object detection.
\newblock In {\em Proceedings of the European conference on computer vision
  (ECCV)}, pages 641--656, 2018.

\bibitem{liang2022bevfusion}
Tingting Liang, Hongwei Xie, Kaicheng Yu, Zhongyu Xia, Zhiwei Lin, Yongtao
  Wang, Tao Tang, Bing Wang, and Zhi Tang.
\newblock Bevfusion: A simple and robust lidar-camera fusion framework.
\newblock {\em Advances in Neural Information Processing Systems},
  35:10421--10434, 2022.

\bibitem{lin2017feature}
Tsung-Yi Lin, Piotr Doll{\'a}r, Ross Girshick, Kaiming He, Bharath Hariharan,
  and Serge Belongie.
\newblock Feature pyramid networks for object detection.
\newblock In {\em Proceedings of the IEEE conference on computer vision and
  pattern recognition}, pages 2117--2125, 2017.

\bibitem{liu2020improving}
Jiang-Jiang Liu, Qibin Hou, Ming-Ming Cheng, Changhu Wang, and Jiashi Feng.
\newblock Improving convolutional networks with self-calibrated convolutions.
\newblock In {\em Proceedings of the IEEE/CVF conference on computer vision and
  pattern recognition}, pages 10096--10105, 2020.

\bibitem{liu2021swin}
Ze Liu, Yutong Lin, Yue Cao, Han Hu, Yixuan Wei, Zheng Zhang, Stephen Lin, and
  Baining Guo.
\newblock Swin transformer: Hierarchical vision transformer using shifted
  windows.
\newblock In {\em Proceedings of the IEEE/CVF international conference on
  computer vision}, pages 10012--10022, 2021.

\bibitem{liu2023bevfusion}
Zhijian Liu, Haotian Tang, Alexander Amini, Xinyu Yang, Huizi Mao, Daniela~L
  Rus, and Song Han.
\newblock Bevfusion: Multi-task multi-sensor fusion with unified bird's-eye
  view representation.
\newblock In {\em 2023 IEEE International Conference on Robotics and Automation
  (ICRA)}, pages 2774--2781. IEEE, 2023.

\bibitem{loshchilov2017decoupled}
Ilya Loshchilov and Frank Hutter.
\newblock Decoupled weight decay regularization.
\newblock {\em arXiv preprint arXiv:1711.05101}, 2017.

\bibitem{man2023bev}
Yunze Man, Liang-Yan Gui, and Yu-Xiong Wang.
\newblock Bev-guided multi-modality fusion for driving perception.
\newblock In {\em Proceedings of the IEEE/CVF Conference on Computer Vision and
  Pattern Recognition}, pages 21960--21969, 2023.

\bibitem{palazzi2017learning}
Andrea Palazzi, Guido Borghi, Davide Abati, Simone Calderara, and Rita
  Cucchiara.
\newblock Learning to map vehicles into bird’s eye view.
\newblock In {\em Image Analysis and Processing-ICIAP 2017: 19th International
  Conference, Catania, Italy, September 11-15, 2017, Proceedings, Part I 19},
  pages 233--243. Springer, 2017.

\bibitem{pan2020cross}
Bowen Pan, Jiankai Sun, Ho~Yin~Tiga Leung, Alex Andonian, and Bolei Zhou.
\newblock Cross-view semantic segmentation for sensing surroundings.
\newblock {\em IEEE Robotics and Automation Letters}, 5(3):4867--4873, 2020.

\bibitem{peng2023bevsegformer}
Lang Peng, Zhirong Chen, Zhangjie Fu, Pengpeng Liang, and Erkang Cheng.
\newblock Bevsegformer: Bird's eye view semantic segmentation from arbitrary
  camera rigs.
\newblock In {\em Proceedings of the IEEE/CVF Winter Conference on Applications
  of Computer Vision}, pages 5935--5943, 2023.

\bibitem{philion2020lift}
Jonah Philion and Sanja Fidler.
\newblock Lift, splat, shoot: Encoding images from arbitrary camera rigs by
  implicitly unprojecting to 3d.
\newblock In {\em Computer Vision--ECCV 2020: 16th European Conference,
  Glasgow, UK, August 23--28, 2020, Proceedings, Part XIV 16}, pages 194--210.
  Springer, 2020.

\bibitem{qiao2023end}
Limeng Qiao, Wenjie Ding, Xi Qiu, and Chi Zhang.
\newblock End-to-end vectorized hd-map construction with piecewise bezier
  curve.
\newblock In {\em Proceedings of the IEEE/CVF Conference on Computer Vision and
  Pattern Recognition}, pages 13218--13228, 2023.

\bibitem{qin2022uniformer}
Zequn Qin, Jingyu Chen, Chao Chen, Xiaozhi Chen, and Xi Li.
\newblock Uniformer: Unified multi-view fusion transformer for spatial-temporal
  representation in bird's-eye-view.
\newblock {\em arXiv preprint arXiv:2207.08536}, 2022.

\bibitem{reiher2020sim2real}
Lennart Reiher, Bastian Lampe, and Lutz Eckstein.
\newblock A sim2real deep learning approach for the transformation of images
  from multiple vehicle-mounted cameras to a semantically segmented image in
  bird’s eye view.
\newblock In {\em 2020 IEEE 23rd International Conference on Intelligent
  Transportation Systems (ITSC)}, pages 1--7. IEEE, 2020.

\bibitem{roddick2018orthographic}
Thomas Roddick, Alex Kendall, and Roberto Cipolla.
\newblock Orthographic feature transform for monocular 3d object detection.
\newblock {\em arXiv preprint arXiv:1811.08188}, 2018.

\bibitem{saha2022translating}
Avishkar Saha, Oscar Mendez, Chris Russell, and Richard Bowden.
\newblock Translating images into maps.
\newblock In {\em 2022 International conference on robotics and automation
  (ICRA)}, pages 9200--9206. IEEE, 2022.

\bibitem{shin2023instagram}
Juyeb Shin, Francois Rameau, Hyeonjun Jeong, and Dongsuk Kum.
\newblock Instagram: Instance-level graph modeling for vectorized hd map
  learning.
\newblock {\em arXiv preprint arXiv:2301.04470}, 2023.

\bibitem{tan2019efficientnet}
Mingxing Tan and Quoc Le.
\newblock Efficientnet: Rethinking model scaling for convolutional neural
  networks.
\newblock In {\em International conference on machine learning}, pages
  6105--6114. PMLR, 2019.

\bibitem{vaswani2017attention}
Ashish Vaswani, Noam Shazeer, Niki Parmar, Jakob Uszkoreit, Llion Jones,
  Aidan~N Gomez, {\L}ukasz Kaiser, and Illia Polosukhin.
\newblock Attention is all you need.
\newblock {\em Advances in neural information processing systems}, 30, 2017.

\bibitem{vora2020pointpainting}
Sourabh Vora, Alex~H Lang, Bassam Helou, and Oscar Beijbom.
\newblock Pointpainting: Sequential fusion for 3d object detection.
\newblock In {\em Proceedings of the IEEE/CVF conference on computer vision and
  pattern recognition}, pages 4604--4612, 2020.

\bibitem{wang2021pointaugmenting}
Chunwei Wang, Chao Ma, Ming Zhu, and Xiaokang Yang.
\newblock Pointaugmenting: Cross-modal augmentation for 3d object detection.
\newblock In {\em Proceedings of the IEEE/CVF Conference on Computer Vision and
  Pattern Recognition}, pages 11794--11803, 2021.

\bibitem{wang2023lidar2map}
Song Wang, Wentong Li, Wenyu Liu, Xiaolu Liu, and Jianke Zhu.
\newblock Lidar2map: In defense of lidar-based semantic map construction using
  online camera distillation.
\newblock In {\em Proceedings of the IEEE/CVF Conference on Computer Vision and
  Pattern Recognition}, pages 5186--5195, 2023.

\bibitem{wang2019pseudo}
Yan Wang, Wei-Lun Chao, Divyansh Garg, Bharath Hariharan, Mark Campbell, and
  Kilian~Q Weinberger.
\newblock Pseudo-lidar from visual depth estimation: Bridging the gap in 3d
  object detection for autonomous driving.
\newblock In {\em Proceedings of the IEEE/CVF Conference on Computer Vision and
  Pattern Recognition}, pages 8445--8453, 2019.

\bibitem{xie2022m}
Enze Xie, Zhiding Yu, Daquan Zhou, Jonah Philion, Anima Anandkumar, Sanja
  Fidler, Ping Luo, and Jose~M Alvarez.
\newblock M$^2$bev: Multi-camera joint 3d detection and segmentation with
  unified birds-eye view representation.
\newblock {\em arXiv preprint arXiv:2204.05088}, 2022.

\bibitem{xie2017aggregated}
Saining Xie, Ross Girshick, Piotr Doll{\'a}r, Zhuowen Tu, and Kaiming He.
\newblock Aggregated residual transformations for deep neural networks.
\newblock In {\em Proceedings of the IEEE conference on computer vision and
  pattern recognition}, pages 1492--1500, 2017.

\bibitem{xu2018pointfusion}
Danfei Xu, Dragomir Anguelov, and Ashesh Jain.
\newblock Pointfusion: Deep sensor fusion for 3d bounding box estimation.
\newblock In {\em Proceedings of the IEEE conference on computer vision and
  pattern recognition}, pages 244--253, 2018.

\bibitem{yang2023bevformer}
Chenyu Yang, Yuntao Chen, Hao Tian, Chenxin Tao, Xizhou Zhu, Zhaoxiang Zhang,
  Gao Huang, Hongyang Li, Yu Qiao, Lewei Lu, et~al.
\newblock Bevformer v2: Adapting modern image backbones to bird's-eye-view
  recognition via perspective supervision.
\newblock In {\em Proceedings of the IEEE/CVF Conference on Computer Vision and
  Pattern Recognition}, pages 17830--17839, 2023.

\bibitem{yang2022deepinteraction}
Zeyu Yang, Jiaqi Chen, Zhenwei Miao, Wei Li, Xiatian Zhu, and Li Zhang.
\newblock Deepinteraction: 3d object detection via modality interaction.
\newblock {\em Advances in Neural Information Processing Systems},
  35:1992--2005, 2022.

\bibitem{yin2021center}
Tianwei Yin, Xingyi Zhou, and Philipp Krahenbuhl.
\newblock Center-based 3d object detection and tracking.
\newblock In {\em Proceedings of the IEEE/CVF conference on computer vision and
  pattern recognition}, pages 11784--11793, 2021.

\bibitem{yin2021multimodal}
Tianwei Yin, Xingyi Zhou, and Philipp Kr{\"a}henb{\"u}hl.
\newblock Multimodal virtual point 3d detection.
\newblock {\em Advances in Neural Information Processing Systems},
  34:16494--16507, 2021.

\bibitem{yu2022metaformer}
Weihao Yu, Mi Luo, Pan Zhou, Chenyang Si, Yichen Zhou, Xinchao Wang, Jiashi
  Feng, and Shuicheng Yan.
\newblock Metaformer is actually what you need for vision.
\newblock In {\em Proceedings of the IEEE/CVF conference on computer vision and
  pattern recognition}, pages 10819--10829, 2022.

\bibitem{zhang2022beverse}
Yunpeng Zhang, Zheng Zhu, Wenzhao Zheng, Junjie Huang, Guan Huang, Jie Zhou,
  and Jiwen Lu.
\newblock Beverse: Unified perception and prediction in birds-eye-view for
  vision-centric autonomous driving.
\newblock {\em arXiv preprint arXiv:2205.09743}, 2022.

\bibitem{zhou2022cross}
Brady Zhou and Philipp Kr{\"a}henb{\"u}hl.
\newblock Cross-view transformers for real-time map-view semantic segmentation.
\newblock In {\em Proceedings of the IEEE/CVF conference on computer vision and
  pattern recognition}, pages 13760--13769, 2022.

\bibitem{zhou2018voxelnet}
Yin Zhou and Oncel Tuzel.
\newblock Voxelnet: End-to-end learning for point cloud based 3d object
  detection.
\newblock In {\em Proceedings of the IEEE conference on computer vision and
  pattern recognition}, pages 4490--4499, 2018.

\bibitem{zhu2021monocular}
Minghan Zhu, Songan Zhang, Yuanxin Zhong, Pingping Lu, Huei Peng, and John
  Lenneman.
\newblock Monocular 3d vehicle detection using uncalibrated traffic cameras
  through homography.
\newblock In {\em 2021 IEEE/RSJ International Conference on Intelligent Robots
  and Systems (IROS)}, pages 3814--3821. IEEE, 2021.

\bibitem{zou2023unim}
Jian Zou, Tianyu Huang, Guanglei Yang, Zhenhua Guo, and Wangmeng Zuo.
\newblock Unim$^{2}$ae: Multi-modal masked autoencoders with unified 3d
  representation for 3d perception in autonomous driving.
\newblock {\em arXiv preprint arXiv:2308.10421}, 2023.

\end{thebibliography}

\end{document}